\pdfoutput=1

\documentclass[11pt]{article}

\usepackage[final]{acl}

\usepackage{times}
\usepackage{latexsym}
\usepackage{amssymb}
\usepackage{hyperref}

\usepackage[T1]{fontenc}

\usepackage[utf8]{inputenc}

\usepackage{microtype}

\usepackage{inconsolata}

\usepackage{listings} 
\lstnewenvironment{prompt}{}{}

\usepackage{multirow}
\usepackage{booktabs}

\newcommand{\comment}[1]{\textcolor{red}{Comment: #1}}

\newcommand{\vname}{forbidden variable}

\newcommand{\actionnoun}{disentanglement}
\newcommand{\actionverb}{disentangle}

\usepackage{graphicx}

\title{Can Large Language Models (or Humans) Disentangle Text?} 

\def\authorsep{\hspace{0.3em}}

\author{Nicolas Audinet de Pieuchon$^{1}$ \authorsep 
Adel Daoud$^{1,2,5}$ \authorsep
Connor T. \ Jerzak$^3$\\
{\bf Moa Johansson}$^1$ \authorsep 
{\bf Richard Johansson}$^{1,4}$ \authorsep 
\\
$^1$Chalmers University of Technology, Sweden \authorsep
$^2$Linköping University, Sweden \\
$^3$University of Texas at Austin, USA \authorsep
$^4$University of Gothenburg, Sweden\\
$^5$The Center for Advanced Studies in Behavioral Sciences, Stanford University, USA. \\
\tt{\{nicolas.audinet, daoud, moa.johansson, richajo\}@chalmers.se},\\
\tt{connor.jerzak@austin.utexas.edu}}

\begin{document}
\maketitle
\begin{abstract}
We investigate the potential of large language models (LLMs) to disentangle text variables---to remove the textual traces of an undesired \vname{} in a task sometimes known as text distillation and closely related to the fairness in AI and causal inference literature. We employ a range of various LLM approaches in an attempt to \actionverb{} text by identifying and removing information about a target variable while preserving other relevant signals. We show that in the strong test of removing sentiment, the statistical association between the processed text and sentiment is still detectable to machine learning classifiers post-LLM-disentanglement. Furthermore, we find that human annotators also struggle to \actionverb{} sentiment while preserving other semantic content. This suggests there may be limited separability between concept variables in some text contexts, highlighting limitations of methods relying on text-level transformations and also raising questions about the robustness of \actionnoun{} methods that achieve statistical independence in representation space.

\end{abstract}

\section{Introduction} 
When computational social scientists analyze text data there are situations where the text is contaminated by a \emph{\vname{}} that we want to preclude from our analysis or handle in a special way. \let\thefootnote\relax\footnotetext{
\noindent To appear as: 
\\ Nicolas Audinet de Pieuchon, Adel Daoud, Connor T. Jerzak, Moa Johansson, Richard Johansson. Can Large Language Models (or Humans) Disentangle Text? {\it Sixth Workshop on NLP and Computational Social Science at NAACL}, 2024.
}
For instance, methods in causal inference that use text to correct for unseen confounders \cite{keith2020text,roberts2020adjusting} run into difficulties if the text is influenced by the treatment variable \cite{daoud2022conceptualizing,gui2023}.
In other contexts, due to ethical, legal, or robustness considerations, we may want to ensure that models trained on a 
corpus are not influenced by effects such as demographic factors \cite{bolukbasi2016man}, domains \cite{ganin2015unsupervised}, personal information \cite{li2018towards}, or other sensitivity information \cite{hovy_five_2021}. 

There are several methods that process text representations to enforce invariance with respect to a \vname{} \cite{barrett2019adversarial,he2020geometric,ravfogel2020null,haghighatkhah2022better,belrose2023leace}.
With a few exceptions, these methods operate on a numerical \emph{representation} of the text, and not directly on the text itself, which makes them less interpretable.
Furthermore, they typically require a large set of annotated examples of the \vname{}, which may not always be available or may be costly to obtain.

Computational social science using text as data is currently being transformed by the introduction of large language models (LLMs) applied in a zero-shot or few-shot fashion \cite{ziems2024llms,tornberg2024best}.
In this paper, we consider the question of whether LLMs have an out-of-the-box ability to \emph{\actionverb{}} a text: to transform it so that the value of a \vname{} is hidden while preserving as much as possible of the original text.
We apply the LLMs in a 
few-shot setup to reproduce the scenario where we have no  large-scale annotations available of the \vname{}. We investigate a variety of prompt-based techniques.
If successful, the \actionnoun{} process would also be understandable to a human reader, since the changes are carried out via interpretable transformations of the text itself.

We find that this type of \actionnoun{} is challenging for the current generation of LLMs, as well as human annotators, to carry out.
While the most powerful LLMs (e.g., GPT-4-class models) sometimes transform text so that it is difficult for humans to determine the original value of the \vname{}, 
we see only a slight reduction in the accuracy of classifiers predicting the \vname{}; its statistical presence is clearly maintained.

\section{Related work}
Our work builds on a growing body of literature on removing undesired information from text representations. This line of research has focused on developing algorithms to learn representations that are independent of protected attributes like demographic variables \citep{li2018towards,raff2018gradient,barrett2019adversarial,belrose2023leace}. These methods typically leverage adversarial training or projection techniques to encourage models to learn representations orthogonal to the \vname{}. 

In addition, some work has also used LLMs for manipulating text, sometimes with goals related to \actionnoun{} in mind. For example, work has used language models for style transfer and controlled text generation, aiming to modify attributes like sentiment or formality while preserving core content \citep{mir2019evaluating,malmi2020unsupervised}. Other research has leveraged language models for tasks like paraphrasing \citep{krishna2020reformulating}, simplification \citep{martin2020controllable}, and neutralizing biased language through LLM fine-tuning \citep{ghanbarzadeh2023gender}. We build on these works by exploring the use of out-of-the-box language models for the task of \vname{} removal via text \actionnoun{}. To the best of our knowledge, we are the first to study the effectiveness of LLM prompting for the \actionnoun{} task systematically and to compare this approach to human performance. 


\section{Defining Disentanglement} 
Previous work has focused on removing a \vname{} from a numerical text representation, not from the raw text itself.
For a representation $X$ and a \vname{} $Z$, 
\newcite{ravfogel2023loglinear} defined \emph{concept erasure} as a process that finds a \emph{guarding function} $h$ such that $h(X) \perp Z$.
%
%
%
Several approaches have been proposed to find guarding functions; 
most work has focused on finding projections that optimize guardedness with respect to linear classifiers \cite{ravfogel2020null,belrose2023leace}.
However, this process is not interpretable in that a human cannot easily reason about what the transformation is doing in the numerical representation space. In addition, finding the guarding function requires a collection of annotated training instances.

In this article, we use a similar conceptualization, but we transform the text directly instead of working in a latent representation space.
We define \emph{text \actionnoun{}} with respect to a \vname{} $Z$ as finding a guarding function $d$ that takes a text and returns a transformed text where textual traces of $Z$ have been removed so that $d(W) \perp Z$. 

We can trivially satisfy this independence criterion by letting $d$ return a transformed text that is unrelated to the original text $W$ (e.g., an empty text). To avoid such uninformative cases, we also want the transformation to be minimally intrusive. In previous work that relied on linear projections, this condition was satisfied by construction.
%

Although the meaning of intrusiveness will vary with research context, we here define it as our ability to measure the effect of \actionnoun{} on our ability to predict other variables represented in the text.
Looking ahead, we envision that minimal intrusiveness can be defined on a semantic level in terms of maximizing similarity in a representation space or on a superficial level by minimizing string edit distance; at this point, we leave this formalization to future work.

\section{Method}
\begin{figure}
    \centering
    \includegraphics[width=0.9\columnwidth]{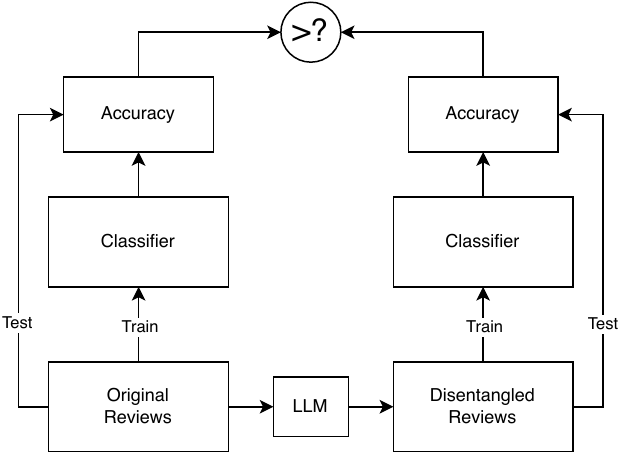}
    \caption{The experimental setup for measuring the effectiveness of LLMs at removing a target variable from the raw text representation.
}
\label{fig:experimental-setup}
\end{figure}

The goal of our experiments is to determine whether current out-of-the-box LLMs are able to \actionverb{} a \vname{} from text without removing traces of other variables. We test two LLMs: Mistral 7B and GPT-4. We choose Mistral 7B as a smaller open-source LLM that performs well on benchmarks and can be run on consumer hardware \cite{Mistral-7B}. We choose GPT-4 as a top-performing commercially available model (as of 2024) \cite{GPT4-technical-report}. More specifically, we use Mistral 7B v0.2 (instruction-tuned) and ``gpt-4-0125-preview'' from the OpenAI API.

To add context to our results, we also test the performance of { mean projection} \cite{haghighatkhah2022better}, a method for removing specific information from representations, as well as, critically, the performance of human annotators. We include human annotators in our experiments to provide a benchmark for the level of \vname{} removal that can be achieved by careful manual text editing, and to explore the intrinsic challenges that may stem from the entanglement of the \vname{} and text content in natural language.

\subsection{Dataset}

In the experiments, we use a dataset consisting of 2000 Amazon reviews. This is a subset of the dataset published by \newcite{amazon_reviews_paper}, which was originally created to investigate domain shifts and domain adaptation of sentiment classifiers. In our subset, each review has a label for the sentiment (positive or negative) and a label for the topic or product category (book, music, camera, health, DVD, or software). The dataset is approximately balanced in both sentiment and topic labels (e.g., 50.65\% of reviews have a positive sentiment; the topic category with the largest number of reviews is ``camera'' with 17.8\%; the category with the smallest number is ``software'' with 15.15\%). We choose the sentiment as the \vname{} and the topic as a proxy for the traces of other variables in the text we wish to keep.

We used this data because, in these reviews, sentiment information tends to be spread throughout the text, rather than localized to a specific sentence or section. This makes sentiment a challenging variable to remove while preserving other text content. As a result, this corpus serves as a strong test case---if LLMs (or humans) can successfully \actionverb{} sentiment from these reviews while retaining the topic information---we would have compelling evidence of their text \actionnoun{} capabilities. In other words, this test is strong in that we expect sentiment information to be spread throughout the text, as opposed to localized in a specific portion of the text, rendering the disentanglement task here more difficult than in the localized case.


\begin{table*}[t]
    \centering
    \begin{tabular}{llcccc}
        \hline
        \textbf{Setting} & \textbf{Prompt} & \textbf{Sentiment Accuracy} $\downarrow$ & \textbf{Topic Accuracy} $\uparrow$ \\
        \hline
        No \actionnoun{} & & 0.885 $\pm$ 0.035 & 0.946 $\pm$ 0.026 \\
        Mean projection & & 0.524 $\pm$ 0.054 & 0.946 $\pm$ 0.026 \\
        \hline
        Human$^*$ & Prompt chaining & 0.800 $\pm$ 0.145 & 0.842 $\pm$ 0.165 \\
        \hline
        \multirow{3}{*}{Mistral 7B} & Paraphrase & 0.891 $\pm$ 0.037 & 0.951 $\pm$ 0.024 \\
        & Few-shot & 0.877 $\pm$ 0.023 & 0.951 $\pm$ 0.015 \\
        & Prompt chaining & 0.841 $\pm$ 0.039 & 0.953 $\pm$ 0.023 \\
        \hline
        \multirow{3}{*}{GPT-4} & Paraphrase & 0.899 $\pm$ 0.034 & 0.951 $\pm$ 0.024 \\
        & Few-shot & 0.824 $\pm$ 0.045 & \textbf{0.955 $\pm$ 0.024} \\
        & Prompt chaining & \textbf{0.757 $\pm$ 0.044} & 0.945 $\pm$ 0.023 \\
    \end{tabular}
    \caption{The impact of disentangling text on sentiment and topic classification accuracy. Results are computed over 2000 Amazon reviews, except for the human setting which was computed on 152 reviews.  $\uparrow$ and $\downarrow$ indicate whether higher values are or are not preferred.}
    \label{tab:experiment-results}
\end{table*}

\subsection{Prompting}

In order to instruct LLMs to \actionverb{} a \vname{} from text, we write the task description as a prompt. Given that the sentiment was chosen as the \vname{}, the LLM is explicitly told to remove the sentiment from the reviews by revising them to be neutral. The LLM is also instructed to keep all other information when revising the reviews. The LLM is not explicitly instructed to keep information about topic, as we want topic to be a proxy for all non-sentiment semantic information.

We investigate two prompting strategies for constructing prompts: few-shot prompting and prompt chaining. Few-shot prompting involves giving one or more examples of how to solve the task as part of the prompt. Doing this has been shown to improve model performance over a zero-shot setup and is one of the most common prompting techniques \cite{few-shot-prompting}. Figure \ref{prompt:few-shot-example} shows an excerpt of the few-shot prompt used in our experiments. The full prompt is shown in Figure \ref{prompt:few-shot} in Appendix \ref{sec:prompts}. 
We provide 3 task examples to the LLM. 

\begin{figure}
    \begin{prompt}[fontsize=\small]
Rewrite the review such that the sentiment is completely neutral. It is very important that one cannot tell whether the review is positive or negative at all. Try and keep all other information in the review.

Here are a few examples of how to do this.

Example 1: [ ... ]; Example 2: [ ... ]; Example 3: [ ... ]

Here's the review: [Review here]
    \end{prompt}
    \caption{Excerpt from the few-shot prompt template. In our tests, [Review here] is replaced with the original text of each review.}
    \label{prompt:few-shot-example}
\end{figure}

We also investigate prompt chaining as an alternative prompting strategy \cite{prompt-chaining}. Prompt chaining involves breaking the problem into smaller tasks and asking the LLM to complete the tasks one at a time, keeping the prompts and answers from previous tasks in context. This strategy can boost performance and gives access to intermediate reasoning steps, which can be used to better understand how the model reasoned about the task.

For our purposes, we use a 2-stage prompt chain. In stage 1, the LLM is asked to return a list of the parts of the text that seem to be associated with the \vname{}. We include 3 examples of how to solve the task as part of the prompt for the first stage. In stage 2, the LLM is asked to rewrite the review from the first stage such that all traces of the \vname{} are removed. Like before, the LLM is also instructed to keep parts of the review not associated with the \vname{}. The full prompts for both stages are available in Figure \ref{prompts:cot-stage-1} and Figure \ref{prompt:cot-stage-2} Appendix \ref{sec:prompts}.

To investigate the default rewriting behavior of the LLMs we also test a {paraphrase} prompt, which asks the model to paraphrase the review without changing the meaning (see Figure \ref{prompt:identity} in Appendix \ref{sec:prompts}). If the LLM perfectly paraphrases the text without altering its semantic content, we would expect the rewritten text to exhibit the same level of association with the \vname{} as the original text. However, if the LLM introduces any changes or artifacts during the rewriting process, this could amplify or diminish the signal of the \vname{} compared to the initial text. 

Finally, we also compare performance with a human baseline. In this baseline, we instructed three people via prompt chaining to (a) list out portions of the text related to sentiment and (b) re-write the text removing those sentiment-related portions but retaining all other information. 

We make the raw reviews, the human-coded sentiment-related content, and the human-disentangled text available at \url{doi.org/10.7910/DVN/TEC1ZP} with replication repository at \url{GitHub.com/AIandGlobalDevelopmentLab/TextDisentanglement}.

\subsection{Evaluation Design}
We use two classifiers to evaluate the effectiveness of LLMs at \actionnoun{}. Both classifiers are trained to label the sentiment of the reviews as this was our chosen \vname{}. The first classifier is trained and tested on the original reviews; the second is trained and tested on the processed reviews. We then compare the accuracy of the two. Given that our dataset is approximately balanced, if the traces of the \vname{} were successfully removed, we would expect the second classifier to have coin-toss accuracy. If traces of the \vname{} were not removed, we would expect the two classifiers to have similar accuracy.

We use the same setup to test whether information about the topic is kept in the processed reviews. See Figure \ref{fig:experimental-setup} for a summary of our evaluation setup.

We use logistic regression trained on document embeddings for all classifiers. The document embeddings are generated by taking the mean of all token embeddings within the document. Token embeddings are generated with DistilBERT \cite{DistilBERT}. To generate confidence intervals we bootstrap over the logistic regressions, using 500 bootstrap samples and a confidence level of 95\%; we use an 80/20 train/test split.



\section{Results}


Our results are summarized in Table \ref{tab:experiment-results}. In general, they show that current out-of-the-box LLMs are unable to consistently remove sentiment from data such as Amazon reviews. Of the LLMs and prompting strategies tested, GPT-4 with prompt chaining achieves best (i.e., lowest) sentiment classifier accuracy with an average of 75.7\% (see an example in Appendix \ref{sec:example}). While the human annotators also struggled with removing traces of the sentiment from the reviews, they achieve results comparable with GPT-4 in sentiment accuracy. The mean projection experiment shows that removing almost all traces of sentiment from the reviews while keeping traces of the other variables is possible when operating at the representation level.

The LLMs performed well at keeping information about topic. Furthermore, the {paraphrase} prompting strategy shows that they successfully keep information about sentiment when rewriting: in the case of GPT-4, the prompt even leads to a slight increase in sentiment classifier accuracy. This indicates that LLMs may amplify original text signals when paraphrasing.

\section{Implications}
While the LLMs generally struggle to remove statistical associations with the \vname{}, we also find that human coders face a similar difficulty. This finding implies that there may be, at least for the task described here, limited separability in the text between the \vname{} and the remainder of the text. 
This limited separability raises questions about the robustness of \actionnoun{} methods operating on downstream representations as such methods may inadvertently generate representations incongruous with real text. Further research is therefore needed to develop techniques that can effectively disentangle the \vname{} from the relevant semantic content in a way that respects the content of the original text.

\section*{Ethical Considerations}
Our work explores the capabilities and limitations of large language models and human annotators in disentangling text variables, raising several important ethical considerations. For example, our findings highlight the challenges of completely removing traces of a target variable while preserving other semantic content in the raw text space. This underscores the importance of transparency and interpretability when applying this or other disentanglement methods, as residual signals may still be detectable even after processing or the processed text representations may be incongruent with the original text semantics. It is important to communicate these limitations to end-users and decision-makers.

\section*{Limitations}

Our experiments focus on variables where the relevant information is spread throughout the text, such as sentiment in product reviews. However, in some applications, the \vname{} may be more localized and separable, such as personal information like names or addresses \cite{hovy_five_2021}. In these cases, the \actionnoun{} task may be easier, as the target information can be more precisely removed.

We also focus on sentiment and topic variables that are relatively independent in our dataset. However, in real-world scenarios, variables of interest may be more intrinsically interrelated, such as political ideology and slant in news articles. Disentangling inherently correlated variables while preserving salient information could be more challenging \cite{daoud2022conceptualizing}. 

Finally, our evaluation results rely on machine learning classifiers, which may not fully capture human perception of the removal of the \vname{}. Classifiers detect statistical patterns but do not necessarily ``read'' text like humans do. There could be cases where classifiers detect residual signals that are not semantically meaningful to humans or where important nuances are lost that are more apparent to humans. To get a more complete picture of \actionnoun{} effectiveness, future work should augment machine evaluations with human judgment experiments, such as having annotators guess the original target variable from the disentangled text.

\section*{Acknowledgements}

This research was supported by the project \emph{Countering Bias in AI Methods in the Social Sciences}
under the 
Wallenberg AI, Autonomous Systems and Software Program -- Humanity and Society (WASP-HS),
funded by the Marianne and Marcus Wallenberg Foundation and the Marcus and Amalia Wallenberg Foundation.

The computations were enabled by resources provided by the National Academic Infrastructure for Supercomputing in Sweden (NAISS) at Alvis partially funded by the Swedish Research Council through grant agreement no.\ 2022-06725.

\bibliography{custom}

\newpage
\appendix

\onecolumn
\section{Appendix}

\subsection{Disentanglement Example}
\label{sec:example}

\begin{figure}[h]
    \begin{lstlisting}
i bought this item because my husband heard is the best internet security . now i regret buying one . since ive installed it , it slowed my computer , its very very hard to close window and worse when i disconnected from the internet i have a hard time logging off . everytime im on the internet , i get mad and upset . its not worth all the stress . i definitely dont recommend this product . now im trying to figure how i can return the product and go back to norton 
    \end{lstlisting}
    \caption{Example of an original review from the Amazon dataset}
    \label{fig:review-example}
\end{figure}

\begin{figure}[h]
    \begin{prompt}
The parts of the review that contain information about the sentiment are:

* my husband heard is the best internet security
* now i regret buying one
* it slowed my computer
* its very very hard to close window
* worse when i disconnected from the internet i have a hard time logging off
* everytime im on the internet, i get mad and upset
* its not worth all the stress
* i definitely dont recommend this product
* now im trying to figure how i can return the product
    \end{prompt}
    \caption{GPT-4 response for the first stage of prompt chaining with the review from Figure \ref{fig:review-example}}
    \label{fig:stage1-example}
\end{figure}

\begin{figure}[h]
    \begin{prompt}
I purchased this item after hearing it was considered a top choice for internet security. Since installing it, I have observed changes in my computer's performance and behavior when disconnecting from the internet, as well as the process required to log off. I am currently exploring options to return the product and considering switching back to Norton.
    \end{prompt}
    \caption{GPT-4 response for the second stage of prompt chaining with the review from Figure \ref{fig:review-example} and the first stage response from Figure \ref{fig:stage1-example}}
\end{figure}

\subsection{Prompts}
\label{sec:prompts}
\begin{figure}[htb]
    \begin{prompt}
Paraphrase the following text without changing the meaning:

[Review here]
    \end{prompt}
    \caption{Prompt template for the ``identity'' experiments.}
    \label{prompt:identity}
\end{figure}

\begin{figure}[htbp]
    \begin{prompt}
Rewrite the review such that the sentiment is completely neutral. It is very important that one cannot tell whether the review is positive or negative at all. Try and keep all other information in the review.

Here are a few examples of how to do this.

Example 1: if the original review was:

i bought this album because i loved the title song . it 's such a great song , how bad can the rest of the album be , right ? well , the rest of the songs are just filler and are n't worth the money i paid for this . it 's either shameless bubblegum or oversentimentalized depressing tripe . kenny chesney is a popular artist and as a result he is in the cookie cutter category of the nashville music scene . he 's gotta pump out the albums so the record company can keep lining their pockets while the suckers out there keep buying this garbage to perpetuate more garbage coming out of that town . i 'll get down off my soapbox now . but country music really needs to get back to it 's roots and stop this pop nonsense . what country music really is and what it is considered to be by mainstream are two different things .

then the neutral rewrite might be:

I bought this album because of the title song. The rest of the album I didn't know as well. Kenny Chesney is a popular artist in the Nashville music scene. He makes many albums with his record company. Country music has been evolving from its roots to a more pop sound.

Example 2: if the original review was:

this is a very good shaver for the private area . however , the key to getting the best results is to trim the longer hairs with scissors or the largest guard first . this will keep the shaver from pulling on the longer hairs and will enable the foil part of the shaver to work . the foil will not be able to do its job if the hairs are too long . the only problem i had with the shaver was that it did not enable me to shave my back like it claimed . however , i use the ' mangroomer ' back shaver for this and it is perfect for you to shave off all your back hair easily with its elongated handle . it is a great product as well . therefore , i would have to say these two products coupled together seem to cover all the bases for men 's grooming on the body . i would highly recommend both of them for perfect manscaping results

then the neutral rewrite might be:

To use this shaver in the private area it is important to trim the longer hairs with scissors or the largest guard first. This will keep the shaver from pulling on the longer hairs and will enable the foil part of the shaver to work. The foil will not be able to do its job if the hairs are too long. The shaver might also not work well on the back. For this, there are other options such as the 'Mangroomer' back shaver which has an elongated handle that makes it easy to shave back hair.

Example 3: if the original review was:

i bought bead fantasies and bead fantasies ii at the same time after reading the positive reviews ; i wish i had looked at these books before buying . there are pretty motifs that i will incorporate into my beading projects but i find the small typed directions overly simplistic and the diagrams are too small . i 'm glad this is n't my first beading book or i would feel totally discouraged from trying any of these projects . i wo n't be buying bead fantasies iii . the art and elegance of beadweaving and coraling technique remain my favorite beading books .

then the neutral rewrite might be:

I bought Bead Fantasies and Bead Fantasies II at the same time. I like some of the motifs but not others. This is not my first beading book. The art and elegance of beadweaving and coraling technique are great beading books. 

Here's the review:

[Review here]
    \end{prompt}
    \caption{Prompt template for the ``few-shot'' experiments.}
    \label{prompt:few-shot}
\end{figure}

\begin{figure}
    \begin{prompt}
Identify the places in the following review which contain information about the sentiment and return them as bullet points.

Here are a few examples of how to do this.

Example 1: if the original review was:

i bought this album because i loved the title song . it 's such a great song , how bad can the rest of the album be , right ? well , the rest of the songs are just filler and are n't worth the money i paid for this . it 's either shameless bubblegum or oversentimentalized depressing tripe . kenny chesney is a popular artist and as a result he is in the cookie cutter category of the nashville music scene . he 's gotta pump out the albums so the record company can keep lining their pockets while the suckers out there keep buying this garbage to perpetuate more garbage coming out of that town . i 'll get down off my soapbox now . but country music really needs to get back to it 's roots and stop this pop nonsense . what country music really is and what it is considered to be by mainstream are two different things .

then the parts of the review that contain information about the sentiment are:

* i loved the title song
* it 's such a great song
* the rest of the songs are just filler and are n't worth the money
* it 's either shameless bubblegum or oversentimentalized depressing tripe
* the suckers out there keep buying this garbage to perpetuate more garbage
coming out of that town
* but country music really needs to get back to it 's roots
* nonsense

Example 2: if the original review was:

this is a very good shaver for the private area . however , the key to getting the best results is to trim the longer hairs with scissors or the largest guard first . this will keep the shaver from pulling on the longer hairs and will enable the foil part of the shaver to work . the foil will not be able to do its job if the hairs are too long . the only problem i had with the shaver was that it did not enable me to shave my back like it claimed . however , i use the ' mangroomer ' back shaver for this and it is perfect for you to shave off all your back hair easily with its elongated handle . it is a great product as well . therefore , i would have to say these two products coupled together seem to cover all the bases for men 's grooming on the body . i would highly recommend both of them for perfect manscaping results

then the parts of the review that contain information about the sentiment are:

* this is a very good shaver for the private area
* the only problem i had with the shaver was that it did not enable me to shave
my back like it claimed
* it is perfect for you to shave off all your back hair easily with its
elongated handle
* it is a great product as well
* i would highly recommend

Example 3: if the original review was:

i bought bead fantasies and bead fantasies ii at the same time after reading the positive reviews ; i wish i had looked at these books before buying . there are pretty motifs that i will incorporate into my beading projects but i find the small typed directions overly simplistic and the diagrams are too small . i 'm glad this is n't my first beading book or i would feel totally discouraged from trying any of these projects . i wo n't be buying bead fantasies iii . the art and elegance of beadweaving and coraling technique remain my favorite beading books .

then the parts of the review that contain information about the sentiment are:

* i wish i had looked at these books before buying
* there are pretty motifs
* i find the small typed directions overly simplistic
* the diagrams are too small
* i 'm glad this is n't my first beading book
* i would feel totally discouraged
* i wo n't be buying bead fantasies iii

Here is the review:

[Review here]
    \end{prompt}
    \caption{Prompt template for the first stage of the ``prompt chaining'' experiments.}
    \label{prompts:cot-stage-1}
\end{figure}

\begin{figure}
    \begin{prompt}
Rewrite the original review such that all the information identified about the sentiment is removed. The goal is to make the review completely neutral. It is very important that one cannot tell whether the review is positive or negative at all. Keep all other information in the review.
    \end{prompt}
    \caption{Prompt template for the second stage of the ``prompt chaining'' experiments.}
    \label{prompt:cot-stage-2}
\end{figure}

\end{document}